\documentclass{midl} % Include author names

\usepackage{mwe} % to get dummy images
\jmlrvolume{-- Under Review}
\jmlryear{2026}
\jmlrworkshop{Full Paper -- MIDL 2026 submission}
\title[Preserving Marker Specificity]{Preserving Marker Specificity with Lightweight Channel-Independent Representation Learning}

\usepackage{booktabs}

\midlauthor{
    \Name{Simon Gutwein\nametag{$^{1,2,3,}$}} \Email{simon.gutwein@ccri.at}\\
    \Name{Arthur Longuefosse\nametag{$^{4}$}} \Email{arthur.longuefosse@riken.jp}\\
    \Name{Jun Seita\nametag{$^{4}$}} \Email{jun.seita@riken.jp}\\
    \Name{Sabine Taschner-Mandl\nametag{$^{1}$}} \Email{sabine.taschner@ccri.at}\\
    \Name{Roxane Licandro\nametag{$^{3,7,8}$}} \Email{roxane.licandro@meduniwien.ac.at}\\
    \addr $^{1}$ St. Anna Children's Cancer Research Institute, Vienna, Austria\\
    \addr $^{2}$ TU Wien, Institute of Visual Computing and Human-Centered Technology, CVL, Vienna, Austria\\
    \addr $^{3}$ Medical University of Vienna, Biomedical Imaging and Image-guided Therapy, Computational Imaging Research, ELIA Group, Vienna, Austria\\
    \addr $^{4}$ RIKEN Center for Integrative Medical Sciences, Medical Data Deep Learning Team, Tokyo, Japan 
    \addr $^{7}$ Medical University of Vienna, Comprehensive Center for AI in Medicine, Vienna, Austria\\
    \addr $^{8}$ Medical University of Vienna, Christian Doppler Lab for Mathematical Modelling and Simulation of Next-Generation Medical Ultrasound Devices, Vienna, Austria
}

\begin{document}

\maketitle
\begin{abstract}
Multiplexed tissue imaging measures dozens of protein markers per cell, yet most deep learning models still apply early channel fusion, assuming shared structure across markers. We investigate whether preserving marker independence, combined with deliberately shallow architectures, provides a more suitable inductive bias for self-supervised representation learning in multiplex data, than increasing model scale. Using a Hodgkin lymphoma CODEX dataset with ~145,000 cells and 49 markers, we compare standard early-fusion CNNs with channel-separated architectures, including a marker-aware baseline and our novel shallow Channel-Independent Model (CIM-S) with 5.5K parameters.
After contrastive pretraining and linear evaluation, early-fusion models show limited ability to retain marker-specific information and struggle particularly with rare-cell discrimination. Channel-independent architectures, and CIM-S in particular, achieve substantially stronger representations despite their compact size. These findings are consistent across multiple self-supervised frameworks, remain stable across augmentation settings, and are reproducible across both the 49 markers and reduced 18 markers settings. These results show that lightweight, channel-independent architectures can match or surpass deep early-fusion CNNs and foundation models for multiplex representation learning. Code is available at \url{https://github.com/SimonBon/CIM-S}
\end{abstract}


\section{Introduction}

Multiplexed biological imaging (MI) technologies such as CODEX~\cite{goltsev2018deep}, MxIF~\cite{gerdes2013highly}, IMC~\cite{giesen2014highly} and MIBI~\cite{angelo2014multiplexed} enable the quantification of a defined set of multiple protein markers at subcellular resolution in biological specimens. These images provide high-dimensional information on cell types, states, and spatial tissue organization. However, their interpretation depends on computational methods that can extract marker-specific spatial features and integrate them into biologically meaningful phenotypes. Existing pipelines rely on cell segmentation~\cite{Windhager2023,Bortolomeazzi2022} followed by aggregation of per-cell marker intensities or handcrafted features. This introduces substantial variability, propagates segmentation errors into downstream analyses~\cite{Bruhns2025,Bai2021}, and limits representation learning to predefined features rather than learning directly from raw multiplex data. Recent advances in computational histopathology and biological imaging increasingly emphasize large-capacity architectures, including foundation models, under the assumption that scale alone improves representation quality~\cite{Matsoukas2024}. While these models have been adapted for multiplex imaging~\cite{Shaban2025,Wang2024}, their designs typically retain early fusion, mixing marker channels from the first layer onward. This implicitly assumes inter-marker correlations that do not generally hold in MI, where each channel represents an independent molecular measurement.\\
An alternative line of work, exemplified by \textit{NeXtMarker} \cite{Gutwein2025}, modifies convolutional architectures to maintain channel-separable computations in early layers, yielding interpretable feature spaces tuned to marker-specific biological structure. This architectural choice aligns with the recent multimodal learning theory showing that early fusion can induce modality collapse \cite{Chaudhuri2025,Wu2024}. When heterogeneous modalities are fused prematurely, predictive signals from weaker modalities become masked, gradients associated with these modalities diminish toward zero, and the model ultimately relies on only a subset of available inputs. In the context of multiplex imaging, such collapse implies that biologically essential markers may cease to contribute to learning, degrading both cell-type discrimination and phenotypic resolution. This challenge is amplified by the contrast with natural image data. In RGB images, channels are strongly correlated, and early fusion is a well-suited inductive bias that captures shared spatial structure \cite{hyperdense}. In multiplexed microscopy, however, channels are statistically independent. Collapsing them at the first layer of a deep architecture mixes chemically distinct signals and can oversmooth the local marker patterns that define cellular phenotypes \cite{shine}. These findings support the view that cellular phenotypes are driven by local stoichiometric patterns rather than deep hierarchical abstractions, meaning that early mixing is the main source of degradation. \\
Motivated by these observations, we hypothesize that preserving marker independence in early layers is a critical architectural inductive bias for multiplex imaging, potentially more important than increasing model scale. We adopt a shallow channel-independent architecture that maintains separable per-marker processing throughout early feature extraction, that enables investigation of how architectural bias influences representation quality, independent of parameter count or model depth. We evaluate our hypothesis in supervised and self-supervised settings using a publicly available CODEX dataset of classical Hodgkin lymphoma, and benchmark against both channel-aware and early-fusion CNNs as well as a recent foundation-scale proteomics model trained on tens of millions of image patches. In addition, we assess whether channel-aware embeddings facilitate interpretable, segmentation-free phenotyping through Layer-wise Relevance Propagation~(LRP)~\cite{Bach2015}. This work's contributions can be summarized as follows:

\begin{enumerate}
    \item Systematic study of the relationship between early channel independence and representational quality in multiplex imaging.
    \item Investigation of architectural inductive bias versus model scale, addressing whether compact architectures can match large pretrained models in this domain.
    \item An evaluation of how architectural design affects marker retention and rare-cell discrimination across full multiplex panels.
    \item A feasibility demonstration of segmentation-free phenotyping enabled by interpretable spatial relevance patterns.
\end{enumerate}

\section{Methods}
\label{sec:method}
Our goal is to evaluate whether preserving marker-wise independence in early layers provides a more suitable inductive bias for multiplex image representation learning than a conventional early-fusion CNN design. To enable a controlled comparison, we construct a deliberately shallow channel-independent architecture in which per-marker structure is retained throughout early feature extraction while keeping overall capacity low.\\

\noindent\textbf{Channel-Independent Model (CIM)}
The CIM encoder architecture processes multiplex images using a channel-independent design. The backbone begins with a grouped convolution where the number of groups equals the number of input markers. This enforces strict per-marker processing: each marker is handled by its own set of filters, with no cross-marker mixing. To provide sufficient representational capacity under this constraint, each marker is expanded into a $k$-dimensional feature space, defined as the width. For each input channel $x_c$, where $c \in \{1,2,\ldots,C\}$ and $C$ denotes the number of markers in the dataset, the layer learns a marker-specific bank of $k$ filters, $\mathcal{W}_c = \{w_{c,1}, \ldots, w_{c,k}\}$, to produce a dedicated feature block:
\[
z_c = x_c * \mathcal{W}_c .
\]
This projection yields $C \times k$ feature maps, organized in such a way, that each marker contributes exactly $k$ independent channels.
Early feature extraction is performed by a stack of lightweight blocks that preserve this channel independence. Each block includes a depthwise $3{\times}3$ convolution for spatial filtering, a squeeze-and-excitation module \cite{hu2018squeeze} that re-weights markers without mixing them, and a grouped $1{\times}1$ convolution to refine features within each marker’s subspace. After these blocks, global average pooling produces a compact representation. All cross-marker fusion is deferred to the final projection or classifier head, ensuring that early layers learn marker-specific structures and enabling a controlled comparison to early-fusion CNNs.\\

\begin{figure}[h!]
    \centering
    \includegraphics[width=\textwidth]{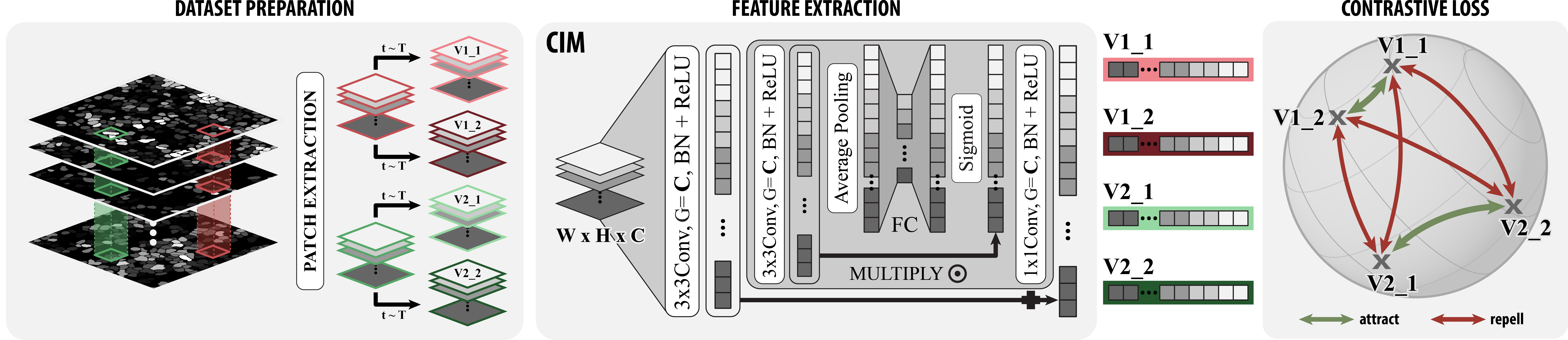}
    \caption{Self-supervised learning pipeline for multiplex single-cell representation learning.  
    \textbf{Left}: Multi-view augmentations (V1, V2) applied to raw patches.  
    \textbf{Middle}: CIM feature extraction with channel-wise grouped convolutions ($G=C$), BN: BatchNom.  
    \textbf{Right}: Contrastive loss driving similarity for positive pairs and dissimilarity for negative pairs.}
    \label{fig:pipeline_ssl}
\end{figure}

\noindent\textbf{Shallow Configuration (CIM-S)}
We adopt a minimal configuration, \textbf{CIM-S} ($N=1$, $k=4$, $\sim$5.5k parameters), to isolate architectural bias from capacity. The architecture is intentionally compact: its purpose is to isolate the role of channel independence rather than to maximize expressivity. CIM-S can be trained in a supervised setting by attaching a classifier head to its feature space, or in a self-supervised setting using approaches such as SimCLR, as illustrated in Fig.~\ref{fig:pipeline_ssl}.

\section{Experiments}
\label{sec:experiments}
In this section, we report two sets of experiments: (i) representation learning, evaluating CIM-S against early-fusion CNNs and recent baselines in supervised and self-supervised settings; and (ii) interpretability and phenotyping, using LRP to assess whether CIM-S supports segmentation-free cell annotation.

\subsection{Dataset}
We use a publicly available CODEX dataset of classical Hodgkin lymphoma (cHL)~\cite{Shaban2024}, containing ~145,000 cells across 49 protein markers. Ground-truth annotations for 16 phenotypes were generated through integrated marker expression analysis with DeepCell segmentation \cite{Greenwald2022}, clustering, and expert curation (Fig.~\ref{fig:dataset}B–C).
Each marker channel is normalized to its 99.9th percentile intensity, and $24 \times 24$ cell-centered patches are extracted as model inputs from the multiplex images.

\subsection{Experiment 1 - Representation Learning Benchmark}
We compare CIM-S against two groups of baselines:
\begin{itemize}
    \item Channel-aware architectures, represented by NeXtMarker \cite{Gutwein2025}, which also preserves marker independence through grouped convolutions.
    \item Early-fusion CNNs, including ResNet18, ResNet50, ResNet101, ResNeXt50-32×4d, and Wide-ResNet50-2. Their first convolution layer is adapted to accept 49 input channels while the remainder of each model is kept unchanged.
\end{itemize}

\noindent Training is carried out under both supervised and self-supervised learning (SSL).\\

\paragraph{Supervised Learning:} Supervised models use a two-layer fully connected classifier and a 70/20/10 train/validation/test split. Data augmentation includes flips, affine transformations, channel-wise intensity scaling, and Gaussian noise. Training is performed for 30 epochs using Adam with a learning rate of $10^{-3}$ and weighted cross-entropy to compensate for class imbalance.\\

\begin{figure}[ht!]
  \centering
  \includegraphics[width=\textwidth]{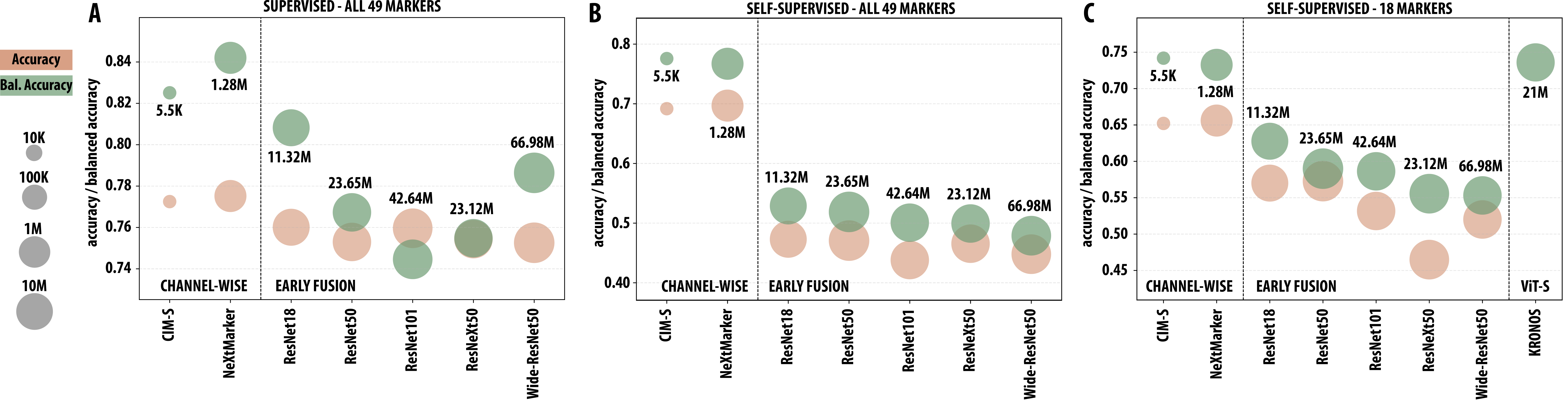}
    \caption{Benchmarking representation quality. A: supervised training on all 49 markers. B: self-supervised training on all 49 markers with linear evaluation. C: self-supervised training on a reduced 18-marker panel, including comparison with the KRONOS foundation model. Bubble size reflects model parameter count; bubble color represents accuracy in orange and balanced accuracy in green.}
  \label{fig:ssl_bubble}
\end{figure}

\paragraph{Results:} As shown in Fig.~\ref{fig:ssl_bubble}A, channel-separable architectures outperform early-fusion CNNs. CIM-S and NeXtMarker reach balanced accuracies of 82.5\% and 84.2\%, respectively. We use balanced accuracy here to account for rare celltypes~\ref{fig:dataset}B. While differences in overall accuracy are smaller, balanced accuracy reveals that channel-aware designs better capture minority phenotypes such as epithelial and mast cells, evident from Fig.~\ref{fig:celltype_report}, which are less prevalent, but clinically relevant.\\

\noindent\textbf{Self-Supervised Learning:} For SSL, we adopt the SimCLR~\cite{Chen2020} framework with temperature~0.2, batch size~512, and the LARS optimizer. Data augmentations are the same as in the supervised setting. CIM-S, NeXtMarker, and ResNet18 are trained for 2000 iterations; larger early-fusion CNNs use 4000 iterations. We additionally evaluate BYOL~\cite{Grill2020} and VICReg~\cite{Bardes2022} to assess robustness across SSL objectives. After pretraining, backbones are frozen and a linear classifier is trained following the supervised protocol. Hyperparameters for SSL pre-training are summarized in Table~\ref{tab:hyperparams}.

\begin{table}[h]
\centering
\caption{Training configurations across architectures. CIM-S achieves competitive performance with orders of magnitude fewer parameters.}
\label{tab:hyperparams}
\begin{tabular}{lcccc}
\toprule
Model & Params (M) & SSL Iterations & Optimizer & LR \\
\midrule
\textbf{CIM-S (Ours)} & \textbf{0.0055} & 2000 & LARS & 0.3 \\
NeXtMarker & 1.28 & 2000 & LARS & 0.3 \\
ResNet18 & 11.3 & 2000 & LARS & 0.3 \\
ResNeXt50-32x4d & 23.1 & 4000 & LARS & 0.3 \\
ResNet50 & 23.6 & 4000 & LARS & 0.3 \\
Wide-ResNet50-2 & 67.0 & 4000 & LARS & 0.3 \\
\bottomrule
\end{tabular}
\end{table}

\paragraph{Results:} Differences between architectures become more pronounced in the self-supervised setting (Fig.~\ref{fig:ssl_bubble}B) and detailed breakdown in Tab.~\ref{tab:ssl}. Early-fusion CNNs trained on all 49 markers achieve balanced accuracies in the 0.45–0.55 range, indicating difficulty in learning meaningful representations without labels, and perhaps modality collapse. In contrast, channel-separable models perform substantially better: CIM-S reaches 0.776, slightly surpassing the larger NeXtMarker (0.765). This reversal of the usual scaling trend suggests that early fusion tends to induce modality collapse in high-dimensional multiplex data, whereas channel-separated processing preserves informative variations across markers.\\

\noindent\textbf{Comparison with foundation models:} Using a reduced 18-marker panel (Fig.~\ref{fig:ssl_bubble}C), CIM-S matches the performance of KRONOS, a foundation-scale model trained on roughly 47 million image patches (74.2\% vs.~73.6\%). This suggests that the right architectural inductive bias can recover discriminatory structure without massive pretraining or large parameter counts.\\

\noindent\textbf{Robustness Across SSL Objectives :} CIM-S performs similarly under SimCLR, BYOL, and VICReg, with SimCLR giving the strongest results (74.2\%). Across all SSL objectives, varying augmentation strength leads to no meaningful performance difference, suggesting that improvements stem primarily from the architectural design rather than sensitivity to specific hyperparameters.\\

\subsection{Experiment 2: Label-Free Phenotyping via Explainability}
\label{subsec:phenotyping}

Accurately annotating cell phenotypes in newly generated datasets is essential for tissue analysis and downstream interpretation. Unlike single-cell transcriptomics, where reference atlases facilitate label transfer, multiplex proteomics suffers from severe batch effects and highly variable marker panels, making direct supervision unreliable~\cite{Stark2020}. Technical differences in staining, acquisition, and image processing further introduce dataset-specific variability. These challenges motivate evaluating whether CIM-S, used as a feature extractor together with LRP, can support label-free cell phenotyping directly from multiplex images.\\

\begin{figure}[ht!]
  \centering
  \includegraphics[width=\textwidth]{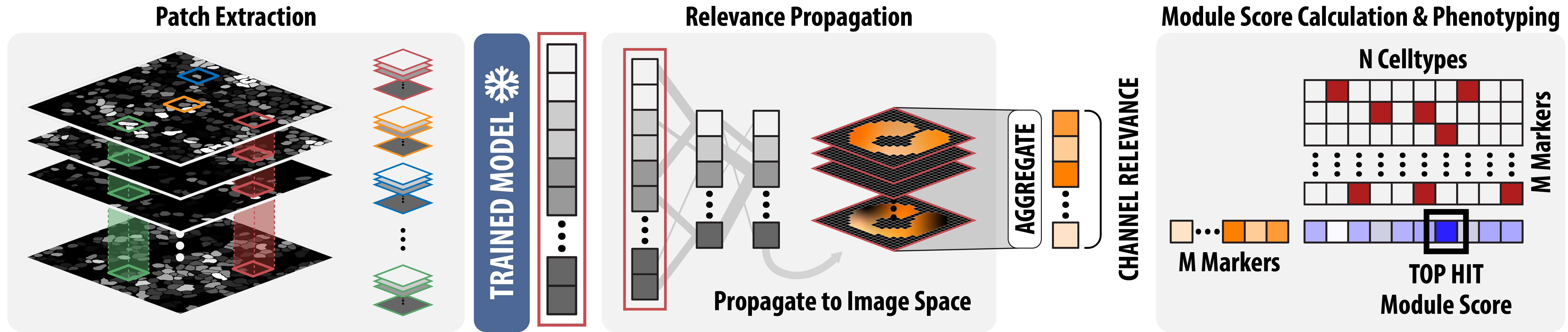}
  \caption{\textbf{Label-free phenotyping workflow.} Raw multiplex images are processed to extract cell-centered patches. A frozen, self-supervised CIM-S encoder computes relevance scores via LRP. These scores are aggregated into marker modules to assign phenotypes without supervised training.}
  \label{fig:phenotyping_workflow}
\end{figure}

\noindent\textbf{Phenotyping Strategy:} For this experiment, we treat the previously used cHL CODEX dataset as a newly acquired, unlabeled dataset without cell-phenotype annotation. The workflow from raw, unsegmented images to phenotyping is summarized in Fig.\ref{fig:phenotyping_workflow}. We define cell types using marker modules based on characteristic co-occurrence patterns described in the literature (Fig.\ref{fig:module_definitions}). We use the CIM-S model trained on the 49-marker setup in Experiment 1, with all weights kept frozen during inference.\\

\noindent\textbf{Relevance Computation:} Cell-level inputs are extracted using the cell probability prior of Cellpose \cite{stringer2021}. We compute spatial LRP maps using the \textit{EpsilonGammaBox} rule \cite{anders2021software}. Spatial maps are aggregated into channel-wise scores by filtering low-magnitude noise, normalizing to the 99\textsuperscript{th} percentile, and weighting by input intensity. Phenotypes are assigned by calculating a module score (average of the top-3 markers of each predefined module, Fig.~\ref{fig:module_definitions}) and selecting the maximum.\\

\begin{figure}[ht!]
  \centering
  \includegraphics[width=0.85\textwidth]{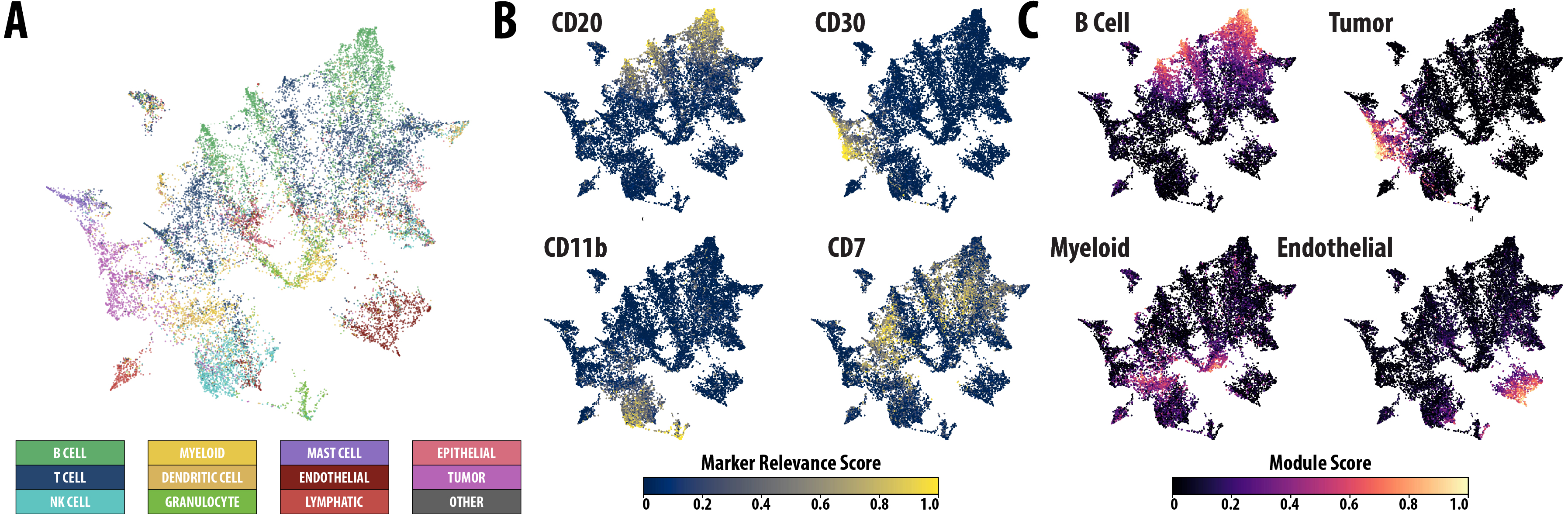}
    \caption{UMAP of the CIM-S feature space for a subset of 20,000 cells from the cHL dataset. (A) Cells colored by the assigned cell type based on the maximum module score. (B) Colored by channel-wise normalized relevance scores for CD20, CD30, CD11b, and CD7. (C) Colored by module score values for the B cell, tumor, myeloid, and endothelial modules.}
  \label{fig:UMAP}
\end{figure}

\noindent\textbf{Results - Latent Space and Efficiency:}
The channel-aware design of CIM-S creates a highly structured latent space even without supervision. Figure~\ref{fig:UMAP}A shows a UMAP visualization colored by assigned cell types. The model forms distinct, concentrated regions corresponding to high relevance for specific markers (Fig.~\ref{fig:UMAP}B), which aligns directly with the module score projections (Fig.~\ref{fig:UMAP}C). \\
A key advantage of CIM-S is its computational efficiency. Due to the model’s minimal memory footprint (approx. 5.5k parameters), the complete end-to-end inference and relevance computation for the entire slide ($\sim$145k patches) requires approximately three minutes on a single NVIDIA V100 32\,GB GPU.\\

\begin{figure}[htb!]
    \centering
    \includegraphics[width=0.9\textwidth]{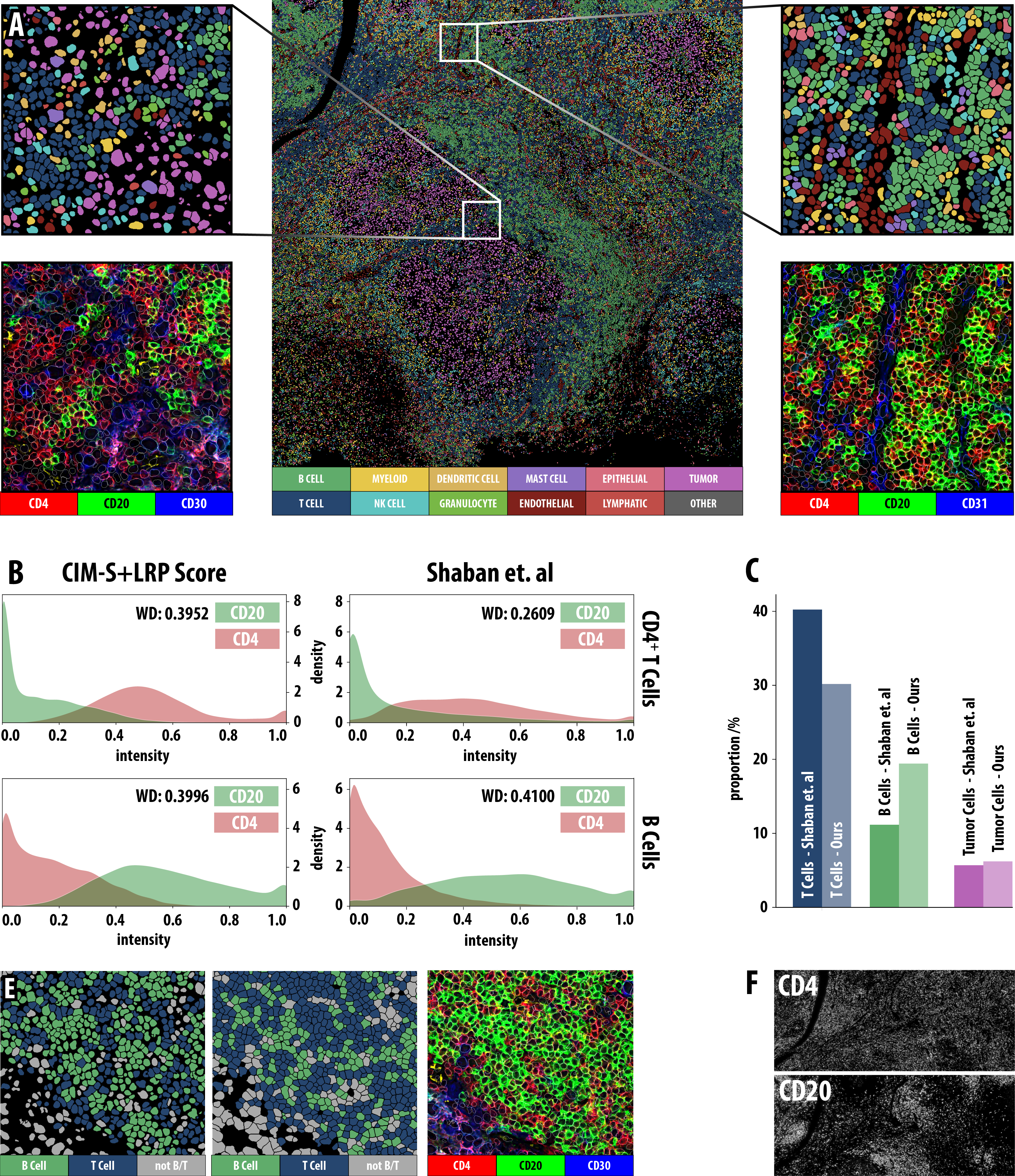}
   \caption{\textbf{Phenotyping Results.} \textbf{A}: Whole-slide annotations and representative local crops illustrating alignment between CIM-S predictions and biological structures (B cells/CD20, T cells/CD4, tumor/CD30, vessels/CD31). \textbf{B}: CD20 vs.\ CD4 distributions for CD4\textsuperscript{+} T cells and B cells, comparing CIM-S LRP relevance scores with raw expression signals from the provided labels. WD: Wasserstein Distance \textbf{C}: Cell-type proportion comparison highlighting the shift in T/B cell ratios. \textbf{E}: Local discordance map between CIM-S (left) and baseline labels. \textbf{F}: Raw signal comparison showing diffuse CD4 background versus sharply localized CD20 signal.}
    \label{fig:phenotyping_results}
\end{figure}

\noindent\textbf{Spatial Analysis and Comparison with Baseline:} Figure~\ref{fig:phenotyping_results}A presents the full-slide phenotyping results. Detailed crops (Fig.~\ref{fig:phenotyping_results}B) confirm that CIM-S annotations align with biological ground truth: tumor cells coincide with CD30 expression, T and B cells align with CD4 and CD20 respectively, and endothelial cells correctly map to CD31 positive (vascular) structures.\\

\noindent\textbf{Discrepancy Analysis:} Comparison with the dataset’s baseline labels reveals a divergence in immune cell proportions: CIM-S assigns 24.8\% fewer T cells and 72.7\% more B cells than the baseline (Fig.~\ref{fig:phenotyping_results}C). This difference is spatially apparent in dense immune regions (Fig.~\ref{fig:phenotyping_results}E). To interpret this result, we emphasize a key biological constraint: CD4 is a canonical T-cell marker and CD20 is a canonical B-cell marker. Under biological definition, these markers are mutually exclusive at the single-cell level; a lymphocyte should express either \textbf{CD4 (T cell)} or \textbf{CD20 (B cell)}, but not both.\\ 
To assess whether the disagreement reflects biological variation or labeling artifacts, we analyzed cells annotated as CD4\textsuperscript{+} T cells and CD20\textsuperscript{+} B cells in the baseline labels and, separately, cells annotated as either group by our phenotyping approach. Within each labeled population, we compared CD4 and CD20 signals using raw intensities and CIM-S relevance scores (Fig.~\ref{fig:phenotyping_results}B). \\
Because CD4 and CD20 should be non-overlapping in an optimal case, we quantified marker separability by computing the Wasserstein distance (WD) between the CD4 and CD20 distributions. For patches annotated as CD4\textsuperscript{+} T cells, the Wasserstein distance is higher for relevance scores than for raw intensities (0.3952 vs.\ 0.2609). This indicates that CIM-S+LRP yields stronger separation between T-cell and B-cell marker information than intensity alone, consistent with hypothesis of baseline misclassification of B cells as T cells due to signal bleed or segmentation ambiguity.\\
Conversely, for cells labeled as B cells, the Wasserstein distance is slightly lower for relevance scores compared to raw intensities (0.3996 vs.\ 0.4100). This suggests that the elevated CD4 signal observed in the B-cell population is not biologically meaningful. Visual inspection (Fig.~\ref{fig:phenotyping_results}F) supports this interpretation: the CD4 raw signal appears diffuse and spatially nonspecific, consistent with background fluorescence, whereas CD20 is sharply localized to cells. CIM-S suppresses this diffuse CD4 signal and assigns T cells primarily where CD20 is absent. Together, these results indicate that the channel-independent architecture provides more specific phenotyping than raw intensity thresholding or early-fusion baselines.

\section{Discussion}
\paragraph{Inductive bias over model scale for multiplex imaging:}
Our results support the hypothesis that preserving marker independence in early feature extraction primarily drives representation quality in multiplex imaging. Across supervised and self-supervised settings, channel-separable networks consistently yield higher balanced accuracy than early-fusion CNNs, indicating improved recovery of minority phenotypes. This pattern aligns with multimodal representation collapse: when heterogeneous inputs are fused early, weaker modalities can become masked and contribute diminishing gradients during training \cite{Chaudhuri2025,Wu2024}. In multiplex imaging, where channels reflect distinct molecular measurements with weak inter-channel correlations and marker-specific noise, early mixing risks entangling informative and confounded signals as marker dimensionality increases. Because cellular phenotypes are determined by local combinations and relative intensities of specific markers, architectures that maintain these stoichiometric patterns for longer appear better suited for this domain. Our results provide empirical evidence that multiplex-specific architectural priors can preserve marker-level signal and improve rare-cell discrimination under realistic panel sizes.\\

\paragraph{Efficient architectures and benchmarking practice:}
The CIM-S and NeXtMarker results show that the benefits of marker-aware inductive bias do not depend on large model capacity. Under the KRONOS 18-marker benchmark, shallow backbones achieve balanced accuracy comparable to a foundation-scale transformer while using orders of magnitude fewer parameters. This underscores the main takeaway of this study: in multiplex imaging, preserving marker independence can substitute for scale. Our label-free analysis also highlights why aggregate classification scores alone may be insufficient:  shifts in inferred B- and T-cell proportions and relevance--intensity discrepancies for CD4 and CD20 suggest that high performance against a single reference labeling can still mask biologically ambiguous or artifact-influenced categories. This aligns with the “gold standard paradox” in digital pathology \cite{Aeffner2017,Rumberger2024}. We therefore recommend that future multiplex benchmarks complement supervised metrics with uncertainty-aware evaluation and attribution plausibility checks.\\

\paragraph{Implications for segmentation-free phenotyping:}
The label-free phenotyping experiment demonstrates that channel-aware embeddings, coupled with LRP, can support segmentation-free cell annotation directly from multiplex images. CIM-S produces spatially coherent phenotypes that align with expected marker--cell type relationships and yields marker relevance profiles that provide interpretable evidence for its assignments. Where our relevance-based annotations diverge from baseline labels, we interpret these cases as candidates for expert review rather than definitive corrections, particularly in regions where non-specific staining or background signal may affect intensity-based pipelines. These results suggest that interpretable, channel-aware models can complement existing segmentation- and intensity-driven workflows and provide a practical route to phenotype discovery when reference labels are incomplete or uncertain.\\

\paragraph{Limitations and future directions:}
This work provides a controlled test of the role of early marker independence in multiplex representation learning and shows consistent gains across supervised and self-supervised settings. Our experiments are currently limited to a single cHL CODEX dataset and one platform, so generalization to other tissues, marker panels, and multiplex technologies remains to be established. The label-free phenotyping results demonstrate feasibility using relevance-guided marker modules; expert or orthogonal validation will be needed for clinical claims. Future work should broaden architectural coverage and evaluate hybrid channel-aware models with large-scale pretraining. The efficiency of CIM-S makes these extensions tractable and supports systematic robustness analysis across batches and acquisition sites.

\clearpage

\bibliography{midl-samplebibliography}
\clearpage

\appendix

\setcounter{figure}{0}
\renewcommand{\thefigure}{S\arabic{figure}}
\renewcommand{\theHfigure}{S\arabic{figure}}

\setcounter{table}{0}
\renewcommand{\thetable}{S\arabic{table}}
\renewcommand{\theHtable}{S\arabic{table}}

\begin{figure}[htb!]
     \centering
    \includegraphics[width=0.99\textwidth]{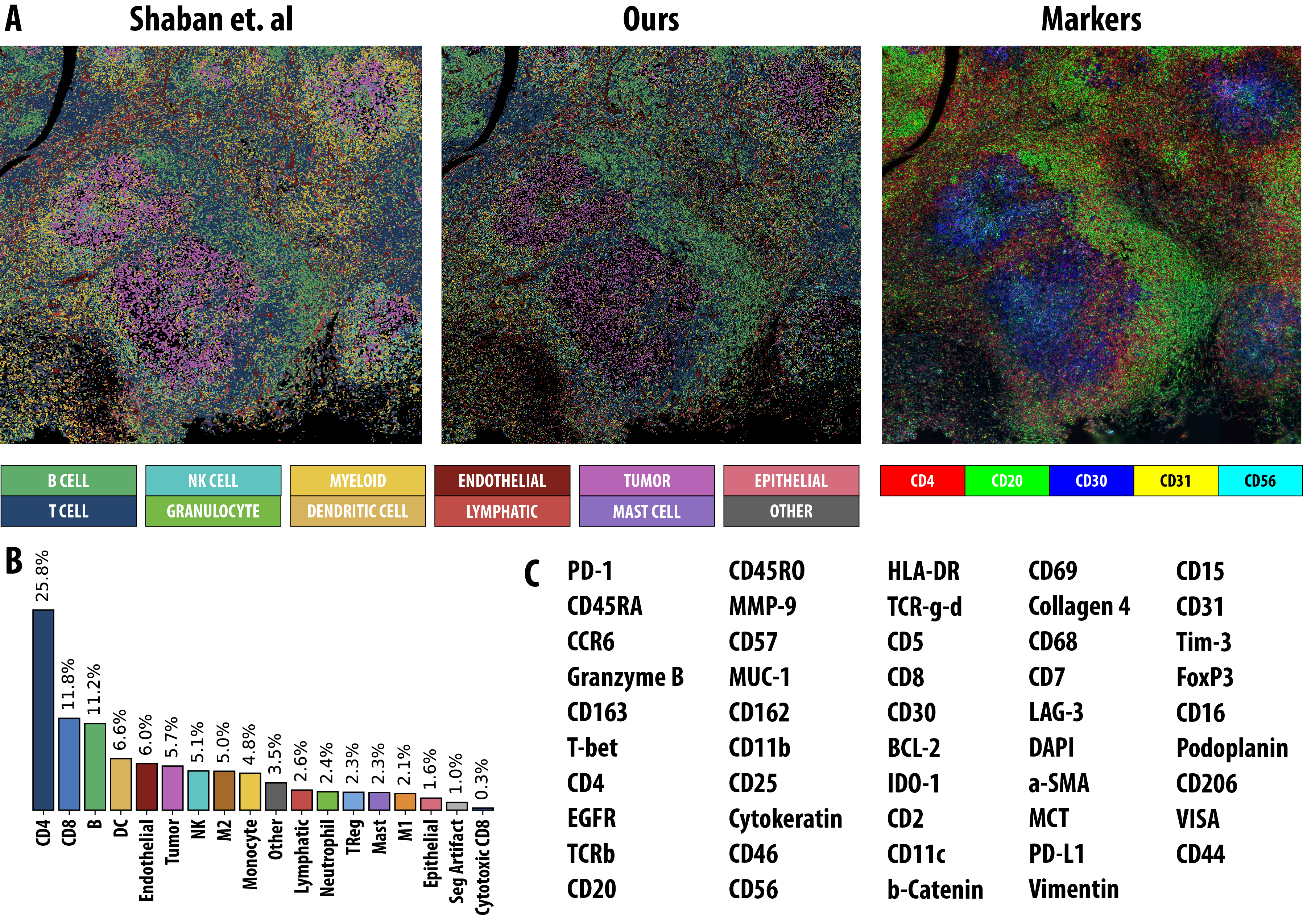}
    \caption{
        \textbf{A}: Cell segmentations colored by the reference labels, our CIM-S annotations, and the corresponding raw marker expression for CD4, CD20, CD30, CD31, and CD56.  
        \textbf{B}: Distribution of annotated cell types (colors match panel A).  
        \textbf{C}: Complete 49-marker panel used in the dataset.
    }
    \label{fig:dataset}
\end{figure}

\begin{figure}[htb!]
\centering
\includegraphics[width=0.9\textwidth]{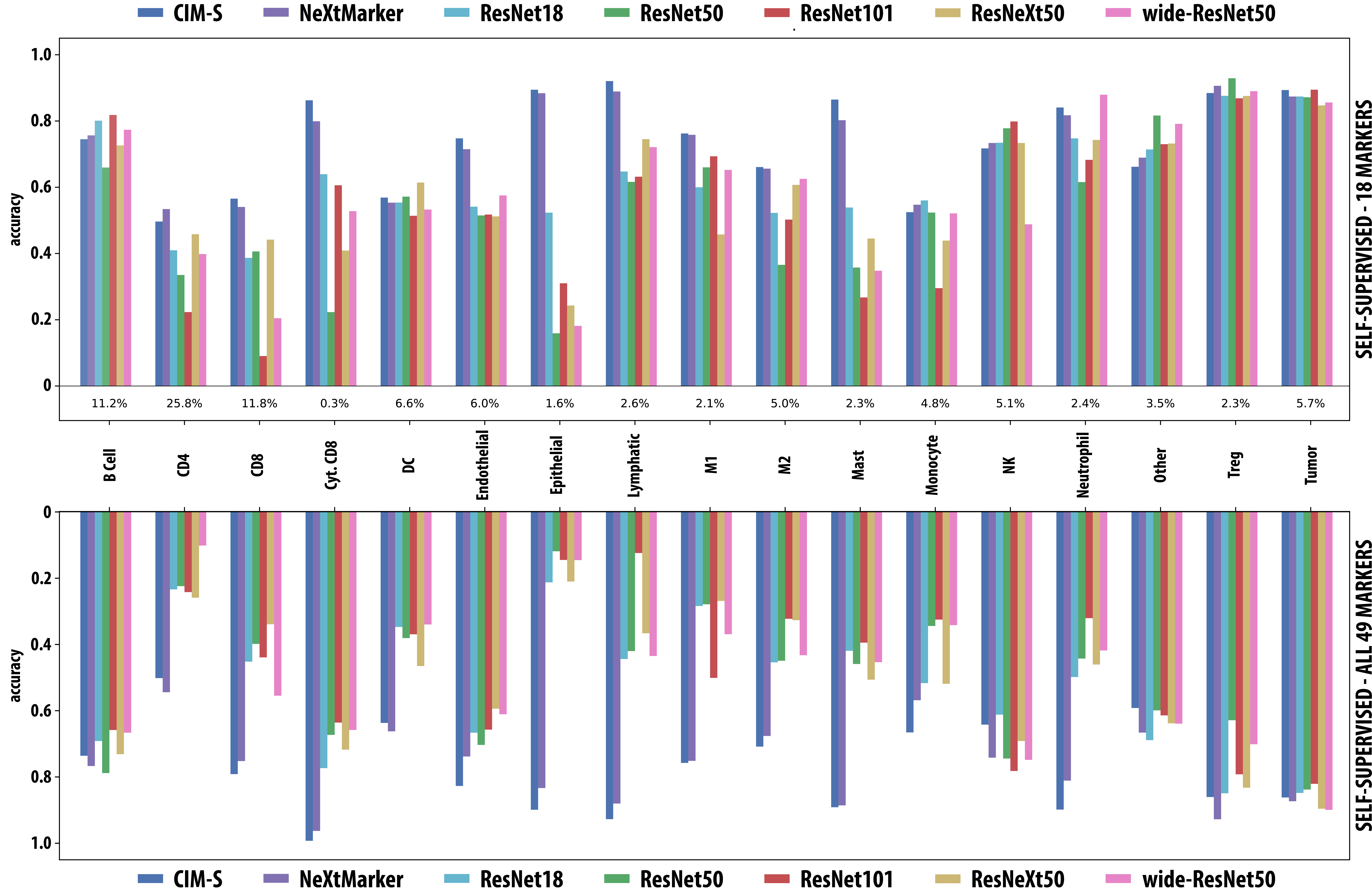}
\caption{Per-cell-type accuracy for CIM-S, NeXtMarker, ResNet-18/50/101, ResNeXt-50, and Wide-ResNet-50 under the self-supervised learning setup. The top and bottom panels report results for the 18-marker and 49-marker panel configurations, respectively.}
\label{fig:celltype_report}
\end{figure}

\begin{table}[htb!]
\small
\centering
\begin{tabular}{lcccccc}
\toprule
\textbf{Model} &
\multicolumn{2}{c}{\textbf{Supervised}} &
\multicolumn{2}{c}{\textbf{SSL 49 markers}} &
\multicolumn{2}{c}{\textbf{SSL 18 markers}} \\
& \textbf{Acc} & \textbf{Bal. Acc}
& \textbf{Acc} & \textbf{Bal. Acc}
& \textbf{Acc} & \textbf{Bal. Acc} \\
\midrule
CIM-S & 77.3 & 82.5 & 69.2 & \textbf{77.6} & 65.2 & \textbf{74.2} \\
NeXtMarker~\cite{Gutwein2025} & \textbf{77.5} & \textbf{84.2} & \textbf{69.7} & 76.7 & \textbf{65.6} & 73.3 \\
\midrule
KRONOS~\cite{Shaban2025} & -- & -- & -- & -- & -- & 73.6 \\
\midrule
resnet18 & 76.0 & 80.8 & 47.3 & 52.9 & 57.0 & 62.8 \\
wide\_resnet50\_2 & 75.3 & 78.6 & 47.1 & 51.9 & 57.3 & 59.0 \\
resnext50\_32x4d & 75.5 & 75.5 & 43.8 & 50.1 & 53.2 & 58.6 \\
resnet101 & 75.9 & 74.5 & 44.8 & 47.9 & 46.5 & 55.6 \\
resnet50 & 75.3 & 76.7 & 46.6 & 49.9 & 52.0 & 55.3 \\
\bottomrule
\end{tabular}
\caption{Performance comparison across architectures and training setups (all values in \%). Best score per column in bold.}
\label{tab:ssl}
\end{table}

\begin{figure}
\centering
\includegraphics[width=0.9\textwidth]{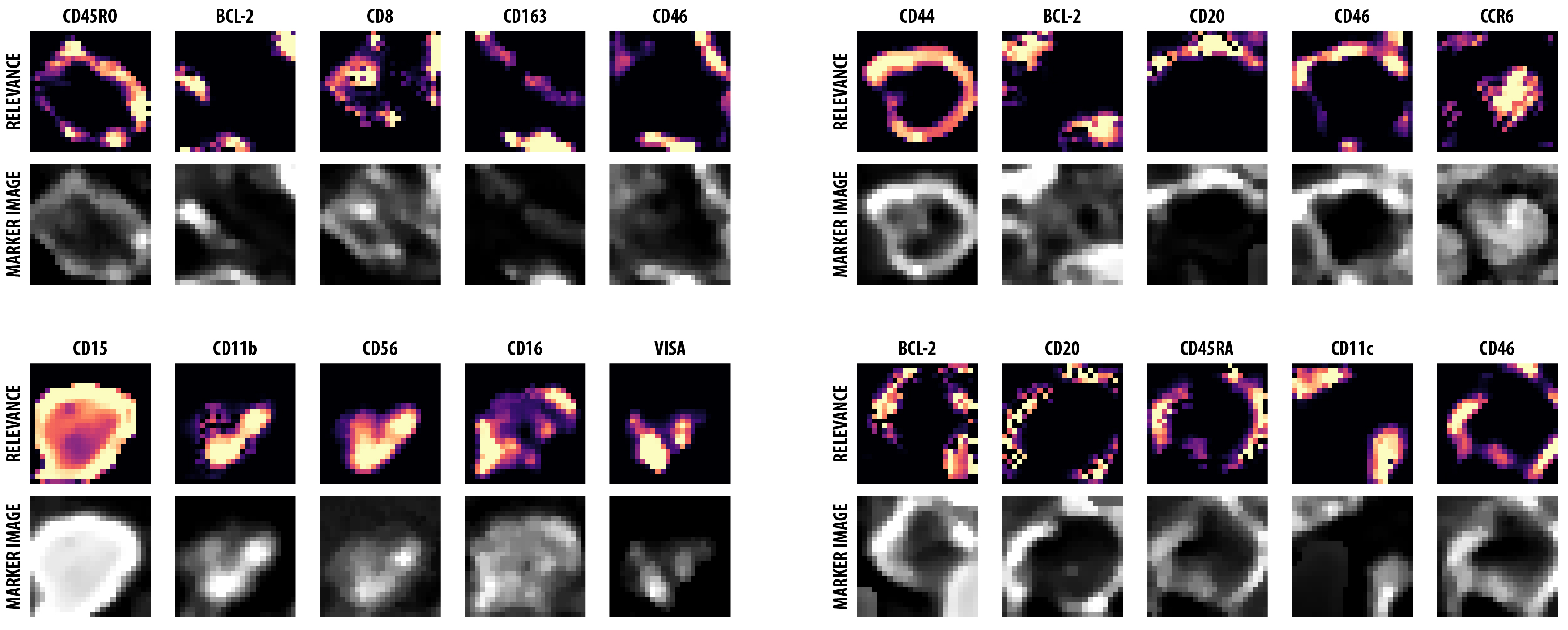}
\caption{Representative LRP visualizations for CIM-S. For each example, the top five spatial relevance channels are shown alongside the corresponding single-cell patch.}
\label{fig:lrp_examples}
\end{figure}

\begin{figure}[t!]
\centering
\includegraphics[width=0.9\textwidth]{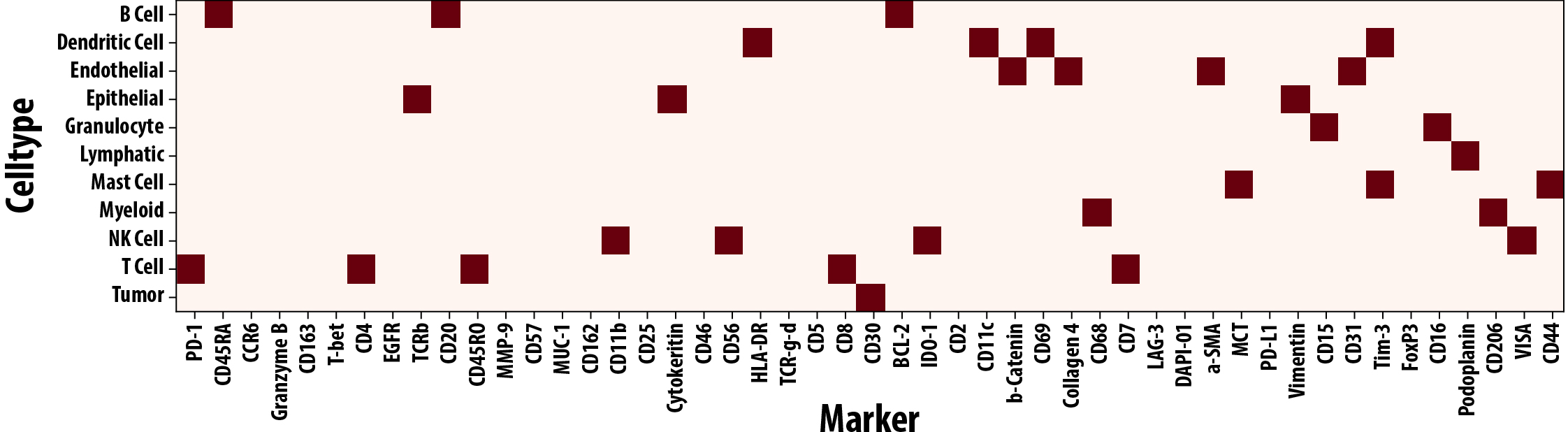}
\caption{Module definitions used for phenotyping. Red circles denote markers included in each cell-type module; all other markers are ignored for that module.}
\label{fig:module_definitions}
\end{figure}

\clearpage

\end{document}